\documentclass[conference]{IEEEtran}

\usepackage{times}
\usepackage{latexsym}
\usepackage[caption=false,font=footnotesize]{subfig}

\usepackage{siunitx}
\usepackage{multirow}  
\usepackage{amsmath}   

\usepackage{tabularx}
\usepackage{graphicx}

\usepackage[T1]{fontenc}

\usepackage[utf8]{inputenc}

\usepackage{microtype}

\usepackage{url}
\usepackage{hyperref}

\usepackage{cite}

\usepackage{listings}
\usepackage{xcolor}

\definecolor{codegray}{gray}{0.95}
\definecolor{codeblue}{rgb}{0.25,0.35,0.75}

\lstdefinestyle{mystyle}{
    backgroundcolor=\color{codegray},
    basicstyle=\ttfamily\footnotesize,
    commentstyle=\color{codeblue}\ttfamily,
    keywordstyle=\bfseries,
    showstringspaces=false,
    breaklines=true,
    frame=single,
    columns=fullflexible,
    literate={_}{{\_}}1, 
}

\usepackage{amssymb}

\usepackage{booktabs}

\usepackage{tikz}
\usetikzlibrary{positioning,arrows}

\usepackage{float}

\begin{document}

\thispagestyle{empty}
\vspace*{1.5in}

\begin{center}
    {\LARGE \textbf{Author Accepted Manuscript}} \\[1.2em]
    {\Large \textbf{Attribution Quality in AI-Generated Content: Benchmarking Style Embeddings and LLM Judges}} \\[1em]
    {\large Misam Abbas} \\[0.3em]
    \textit{Accepted for presentation at the IEEE International Conference on Data Mining Workshops (ICDM  Workshop on Grounding Documents with Reasoning, Agents, Retrieval, and Attribution 2025)
)} \\[2em]
\end{center}

\vfill

\noindent
© 2025 IEEE. Personal use of this material is permitted. Permission from IEEE must be obtained for all other uses, in any current or future media, including reprinting/republishing this material for advertising or promotional purposes, creating new collective works, for resale or redistribution to servers or lists, or reuse of any copyrighted component of this work in other works. This is the author’s accepted version of the paper accepted for presentation at IEEE ICDM Workshops 2025. The final published version will be available at IEEE Conference Workshop proceedings. \\[1.5em]

\noindent
\textbf{Citation:}  
M. Abbas, “Attribution Quality in AI-Generated Content: Benchmarking Style Embeddings and LLM Judges,” in *Proc. IEEE ICDM Workshops 2025* (to appear). \\[1em]

\noindent
\textbf{BibTeX:}
\begin{verbatim}
@inproceedings{abbas2025attribution,
  author    = {Misam Abbas},
  title     = {Attribution Quality in AI-Generated Content:
               Benchmarking Style Embeddings and LLM Judges},
  booktitle = {Proceedings of IEEE ICDM Workshops (ICDMW)},
  year      = {2025},
  note      = {Author Accepted Manuscript}
}
\end{verbatim}

\vspace{1in}

\clearpage

\title{Attribution Quality in AI-Generated Content: Benchmarking Style Embeddings and LLM Judges}

\author{\IEEEauthorblockN{Misam Abbas, \textit{Senior Member, IEEE}}
\IEEEauthorblockA{West Windsor, NJ, USA \\
misam.abbas@gmail.com}
}

\maketitle

\IEEEpubid{\begin{minipage}{\textwidth}\centering
© 2025 IEEE. Personal use of this material is permitted. Permission from IEEE must be obtained for all other uses, in any current or future media, including reprinting/republishing this material for advertising or promotional purposes, creating new collective works, for resale or redistribution to servers or lists, or reuse of any copyrighted component of this work in other works. This is the author’s accepted version of the paper accepted for presentation at IEEE ICDM Workshops 2025. The final published version will be available at IEEE Xplore.
\end{minipage}}

\begin{abstract}
Attributing authorship in the era of large language models (LLMs) is increasingly challenging as machine-generated prose rivals human writing. 
We benchmark two complementary attribution mechanisms—fixed \emph{Style Embeddings} and an instruction-tuned LLM judge (\textsc{GPT-4o})—on the \textsc{Human–AI Parallel  Corpus}, an open dataset from which we choose 600 balanced instances spanning six domains (academic, news, fiction, blogs, spoken transcripts, and TV/movie scripts). Each instance contains a human prompt with both a gold continuation and an LLM-generated continuation from either \textsc{GPT-4o} or \textsc{LLaMA-70B-Instruct}. The Style Embedding baseline achieves stronger aggregate accuracy on GPT continuations (82\% vs.\ 68\%). The LLM Judge is slightly better than the Style embeddings on LLaMA continuations (85\% vs.\ 81\%) but the results are not statistically significant.
Crucially, the LLM judge significantly outperforms in \emph{fiction} and \emph{academic} prose, indicating semantic sensitivity, whereas embeddings dominate in \emph{spoken} and \emph{scripted dialogue}, reflecting structural strengths. These complementary patterns highlight attribution as a multidimensional problem requiring hybrid strategies. 
To support reproducibility we provide code in GitHub repositories and derived data on Huggingface. We release both the dataset and source code under the MIT license. 
This open framework provides a reproducible benchmark for attribution quality assessment in AI-generated content. We also provide a thorough review of prior literature particularly the papers that directly influenced the framework and results in this paper.

\end{abstract}


\section{Introduction}

Large language models (LLMs) now generate prose that often passes casual
human inspection, blurring the boundary between human and machine
authorship \cite{bhandarkar2024emulating,reinhart2024llms}. Reliable
provenance detection is therefore essential for tasks ranging from
academic integrity \cite{cotton2023chatting} to content moderation \cite{kreps2022news}. Prior work approaches this
problem from two directions: (i) fixed encoders that model stylistic
regularities and register‐specific markers
\cite{wegmann2022same}, and (ii) LLMs themselves, prompted to act as
judges of text authenticity \cite{shi2025impersona}. Yet the relative
strengths of these paradigms remain unclear, especially across diverse
genres and generator families.

To investigate these attribution mechanisms systematically, we frame the
problem within a controlled benchmark and evaluation setting. Using the
\textsc{Human–AI Parallel Corpus} as a balanced testbed, we design an
evaluation protocol that pairs human and LLM-generated continuations
across diverse domains. Accuracy serves as the primary attribution
metric, complemented by McNemar’s test to assess whether differences
between methods reflect systematic trends rather than random variation.
In this way, our hypothesis testing doubles as a demonstration of how this
benchmark and evaluation framework can be structured to assess
attribution quality with statistical rigor.

Works such as \cite{zheng2023judging} has highlighted both the promise and the challenges of using LLMs themselves as evaluation judges, reinforcing the importance of systematic benchmarks like ours to quantify attribution quality across domains.

\textbf{Hypothesis.}  We test the hypothesis that \emph{instruction-tuned
LLMs \cite{wei2022finetuned} can reliably discriminate between human-authored and LLM-generated
continuations of a text}. To gauge the limits of this capability, we
contrast LLM judges with a strong non-parametric baseline derived from
style embeddings \cite{wegmann2022same}. Our expectation is that if
LLMs possess an internal representation of their own generative
footprint, then, prompted appropriately, they should at least match, and
possibly exceed, the performance of a stylistic similarity heuristic.

Our primary metric is accuracy, justified by the balanced 50/50 class
distribution; random chance is 0.50. To determine whether observed
differences are statistically significant, we apply McNemar’s test \cite{dietterich1998approximate} to
paired predictions. Because both systems (LLM judge and style embeddings) predict labels on the same set of examples with ground-truth available, we use McNemar’s test—a statistical test on paired nominal data — to assess whether their accuracy differences are statistically significant within each domain. 

Our results show that while the style-embedding baseline attains higher aggregate accuracy, the LLM judge surpasses it
in fiction and academic prose—domains where deeper semantic and
narrative coherence may be decisive. These findings nuance the original
hypothesis, revealing both the promise and the current limitations of
LLMs as self-auditors of textual authenticity.

\section{Related Work}

\subsection{IMPersona: Evaluating Individual Level LM Impersonation}

This paper \cite{shi2025impersona}  investigates the capability of LLMs to impersonate specific individuals given the history of their text messages. This task has two aspects, mimicking the style as well as storing and retrieving information in context. The paper devises a novel setup for evaluating the impersonation capabilities of the model. Essentially it uses personal text conversations of participants to set up LLMs (through fine-tuning and in-context prompts) to impersonate those participants and have conversations with people known to them. They then measure the success rate of the LLMs to fool the acquaintances. For the impersonation to be effective the impersonation includes both stylistic imitation and context fidelity.The participants in the study were given guidance on the topics to discuss. These included stylistic topics where contextual information is not needed, examples such as: "Which furniture would have the most interesting
stories" , "Whether a hot dog is a sandwich" and contextual topics like "Song or album that puts you in a good mood"

\subsubsection{Prompt Used for Impersonation}

Here is an excerpt from the prompt that they used:

\textit{"Today's date is 2025-03-23. You are a human being named
[REDACTED]. You are not an AI. Respond as yourself. You will first be
given the topic of conversation, then any existing conversation history
if there is any. Be sure to reply in the style of [REDACTED]. Use the
'<|msg|>' token to send multiple messages at once if you wish.
Here is an example conversation between [REDACTED] and another individual.
This conversation is not relevant to the current conversation. Use this
conversation to help aid you to emulate stylistically on how to ..."}

\subsubsection{Conclusions from the paper}

\paragraph{}
Their overall conclusions are the following
\begin{itemize}
    \item In context learning for style imitations does not work well even for large models.
    \item Fine-tuning approach works better but it sometimes develops flow problems e.g. jumping between topics.
    \item Still, it is possible to impersonate individuals with limited data which is concerning.
    \item They note that texting frequency between two participants and previous AI use were the two best predictors of whether the imitation would be detected.
\end{itemize}

\subsection{Same Author or Just Same Topic? Towards Content-Independent Style Representations}

This paper  {\cite{wegmann2022same}}  discusses when learned style representations are used for detecting stylistic information for authorship verification tasks, a key issue is that the works or authors are often about the same or similar topics so the learned representations conflate topic with style. 
The researchers propose two innovations: 
(1) a Contrastive Authorship Verification (CAV) setup that adds a "contrastive" utterance from a different author, and 
(2) content control (CC) using conversation or domain labels as topic proxies.
 
Using Reddit data with utterances from 100 subreddits, they train siamese BERT-based networks \cite{bromley1993signature} with different content control methods (conversation, domain, no control) and evaluate using the STEL framework. 
The conversation control works like this, 3 sentences A1, A2 and B are selected where A1 and A2 are written by the same author but in different conversations but B is in the same conversation as A1 but written by a different author. The purpose of the task setups is to embed A1 and A2 closer than A1 and B. 
They find that style embeddings trained with setup as above represent style in a way that is more independent from content than earlier methods. They also demonstrate the elements of style captured by the embeddings through agglomerative clustering of the generated embeddings and identifying style information in those clusters.

\subsection{Emulating Author Style: A Feasibility Study of Prompt-enabled Text Stylization with Off-the-Shelf LLMs }

This paper \cite{bhandarkar2024emulating} explores whether off-the-shelf LLMs can accurately emulate an author's writing style through in context learning.

\subsubsection{Prompt used for Text Stylization}
 
\textbf{Task:}

\textit{<SYS> You are an emulator designed to
replicate the writing style of a human author.<\
SYS> Your task is to generate a 500-word continuation that seamlessly integrates with the provided human-authored snippet. Strive to make the continuation indistinguishable from the human-authored text.}

\textbf{Instructions:}

\textit{The goal of this task is to mimic the author’s writing style while paying meticulous attention to lexical richness and diversity,
sentence structure, punctuation style, special character style, expressions and idioms, overall tone, emotion and mood, or any other relevant aspect of writing style established by the author.}

\textbf{Output Indicator:}

\textit{As output, exclusively return the text completion without any accompanying explanations or comments.}
\textbf{Example author text}

\textit{Text snippet : [50 or 300-word human authored text]}

\subsubsection{Description of Task setup}

The research paper evaluates 12 pre-trained LLMs on author style emulation using guided instructions. They implemented a prompting protocol with two main approaches:

(1) Trivial Emulation Protocol (TEP), providing a simple task definition with a short author text snippet (2) Complex Emulation Protocol (CEP), providing additional author data and guided instructions with linguistic features.

Overall the authors conclude that the performance of current LLMs was underwhelming for prompt based style emulation. For future work the authors suggest exploring more sophisticated prompting techniques.

\subsection{Do LLMs write like humans? Variation in grammatical and rhetorical styles }

This paper \cite{reinhart2024llms} explores whether large language models truly write like humans beyond basic grammar and vocabulary. While LLMs have demonstrated impressive reasoning capabilities, their linguistic performance has received less systematic analysis. The researchers investigate grammatical, lexical, and rhetorical differences between human and LLM writing. Researchers developed two parallel corpora, COCA AI Parallel (CAP) and Human AI Parallel English (HAP-E), consisting of human texts and LLM-generated continuations.

\subsubsection{Prompt used for style emulation}

Instruction-tuned LLMs were given the following prompt, followed by the first chunk of human text:

\textit{"In the same style, tone, and diction of the following text, complete the next 500 words, generate exactly 500 words, and note that the text does not necessarily end after the generated words:"}

\subsubsection{Evaluations and results}

Six LLMs were evaluated including GPT-4o, GPT-4o Mini, and 3 Llama variants. The LLMs were instructed to produce 500-word continuations from 500-word human prompts. They analysed 66 linguistic features from Douglas Biber's tagset \cite{biber1988variation}, encompassing grammatical structures, rhetorical patterns, and lexical choices. Random forest and lasso-penalized logistic regression models were used to classify sources and identify distinguishing features. The 7 class (1 human, 6 LLMs) random forest classifier achieved 66 percent accuracy (vs. 14 percent random guessing) in distinguishing between human and six LLM sources. Instruction-tuned LLMs used present participial clauses 2-5 times more frequently than humans and nominalizations 1.5-2 times more often. GPT-4o and GPT-4o Mini overused words like "camaraderie" and "palpable" at greater than 100 times human rates. Interestingly the performance of LLama Base models in emulating human text was better than the performance of instruction tuned variants.

\section{Methodology}

\subsection{Task Formulation}

\textbf{Task formulation.}  From the \textsc{Human–AI Parallel Corpus}
\cite{reinhart2024llms}, we sample 600 instances spanning six domains
(\textit{academic}, \textit{news}, \textit{fiction}, \textit{blogs},
\textit{TV/movie scripts}, and \textit{spoken}). Each instance contains
a 500-word prompt (T\textsubscript{1}) and two 500-word continuations:
the gold human continuation and a continuation produced by either
\textsc{GPT-4o} or \textsc{LLaMA-70B-Instruct}. The classifier must
identify the human continuation.

We define a binary classification task to assess a model’s ability to distinguish between human-written and machine-generated continuations of a text passage. Each instance is constructed as follows:

\begin{itemize}
    \item \textbf{T\textsubscript{1}}: A 500-word excerpt from an original text.
    \item \textbf{Chunk 2}: The human-authored continuation of T\textsubscript{1} (also 500 words).
    \item \textbf{LLM continuation}: A 500-word continuation generated by a large language model (either \textsc{LLaMA-70B-Instruct} or \textsc{GPT-4o}).
\end{itemize}

For each instance, we randomly assign the two continuations (Chunk 2 and the LLM-generated text) to positions A and B. The task is to determine which of the two candidates is the true human-written continuation of T\textsubscript{1}. 

\begin{figure}[H]
\centering
\begin{tikzpicture}[node distance=1.5cm]
    \node[draw, rectangle] (input) {T1: Original Text};
    \node[draw, rectangle, below left of=input] (optA) {Option A};
    \node[draw, rectangle, below right of=input] (optB) {Option B};
    \node[draw, rectangle, below=2cm of input] (judge) {Binary Classification};
    \node[draw, rectangle, below of=judge] (output) {Which is human continuation- A or B?};
    
    \draw[->] (input) -- (optA);
    \draw[->] (input) -- (optB);
    \draw[->] (optA) -- (judge);
    \draw[->] (optB) -- (judge);
    \draw[->] (judge) -- (output);
\end{tikzpicture}
\caption{Task Formulation}
\label{fig:task}
\end{figure}
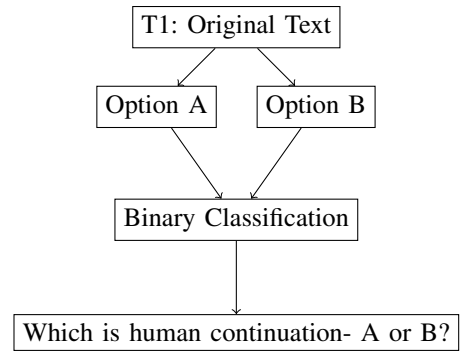

\subsection{Data}

We used a subset of the Human-AI parallel corpus here: 
{Do LLMs write like humans.." \cite{reinhart2024llms}} This dataset is available via huggingface here : \url{https://huggingface.co/datasets/browndw/human-ai-parallel-corpus}

We used 100 random samples picked from each of the following 6 different source types. 

\begin{enumerate}
    \item Academic: 40,000+ open-access Elsevier academic articles
    \item News: 100,000+ online articles from U.S. news organizations
    \item Fiction: Public domain novels and short stories from Project Gutenberg
    \item Spoken: 100,000 podcast transcriptions (differs from COCA's unscripted news show transcripts)
    \item Blogs: 681,288 posts from blogger.com
    \item TV/Movie Scripts: Combined from two script corpora, some OCR-converted
\end{enumerate}

\subsubsection{Dataset Description}

\begin{itemize}
    \item The dataset is designed for comparing human-authored writing with LLM-generated text, containing parallel 500-word chunks where LLMs continue text from an initial human-written segment.
    \item It spans 6 diverse text types: academic articles, news, fiction, spoken (podcast transcriptions), blogs, and TV/movie scripts, with approximately 8,290 texts for each text type and author.
    \item Its creation methodology involved seeding LLMs with human-authored 500-word chunks and prompting them to produce another 500 words, allowing direct comparison with what human authors actually wrote next.
    \item Instruction-tuned LLMs were given the following
prompt, followed by the first chunk of human text:
"In the same style, tone, and diction of the following text, 
complete the next 500 words, generate
exactly 500 words, and note that the text does not
necessarily end after the generated words:" 
\end{itemize}

\subsection{Model}

\subsubsection{Models for Comparison}

The following models are compared in the experiments that follow:

\begin{enumerate}
    \item \textbf{Style-Embedding Baseline}: we reuse the encoder of
    \cite{wegmann2022same} and choose the continuation with higher
    cosine similarity to T\textsubscript{1}.
    \item \textbf{LLM Judge}: we issue a zero-shot binary prompt to
    \textsc{GPT-4o} \cite{openai2024gpt4} asking which continuation is human-written.
\end{enumerate}

\subsubsection{Baseline Model - Style Embeddings baseline}

The Baseline model is illustrated in Fig.2

\begin{figure}[H]
\centering
\begin{tikzpicture}[node distance=1.5cm]
    \node[draw, rectangle] (input) {T1: Original Text};
    \node[draw, rectangle, below left of=input] (optA) {Option A};
    \node[draw, rectangle, below right of=input] (optB) {Option B};
    \node[draw, rectangle, below=2cm of input] (judge) {Cosine Similarities};
    \node[draw, rectangle, below of=judge] (output) { Which is greater cossim(T1,A) or sim (T1,B) ?};
    
    \draw[->] (input) -- (optA);
    \draw[->] (input) -- (optB);
    \draw[->] (optA) -- (judge);
    \draw[->] (optB) -- (judge);
    \draw[->] (judge) -- (output);
\end{tikzpicture}
\caption{Style Embeddings. Binary human vs. LLM continuation detection.}
\label{fig:Baseline}
\end{figure}
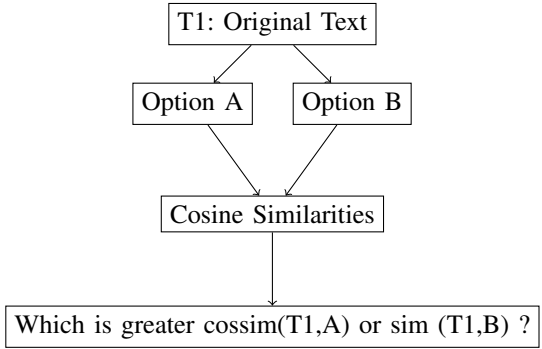

These embeddings are made available by the paper {Same Author or Just Same Topic..." \cite{wegmann2022same}}. 

\url{https://huggingface.co/AnnaWegmann/Style-Embedding}

To establish a baseline, we compute style embeddings for the original text chunk and its two candidate continuations. We then predict the continuation whose embedding exhibits a higher cosine similarity \cite{mikolov2013efficient} with that of the original chunk.

\paragraph{Embedding-Based Prediction.} We compute style embeddings for T\textsubscript{1}, A, and B using a pretrained encoder. The predicted continuation is the candidate with the higher cosine similarity to the original text:
\begin{equation}
\label{eq:prediction}
\text{Predict A if } \cos(\text{T\textsubscript{1}}, A) > \cos(\text{T\textsubscript{1}}, B); \text{ else B}.
\end{equation}

We follow the notebook made available by the authors of the paper from the GitHub repository above.

\subsubsection{LLM-as-a-Judge Model}

The LLM as a Judge Model is illustrated in Fig. 3

\begin{figure}[H]
\centering
\begin{tikzpicture}[node distance=1.5cm]
    \node[draw, rectangle] (input) {T1: Original Text};
    \node[draw, rectangle, below left of=input] (optA) {Option A};
    \node[draw, rectangle, below right of=input] (optB) {Option B};
    \node[draw, rectangle, below=2cm of input] (judge) {LLM as a Judge};
    \node[draw, rectangle, below of=judge] (output) {Which one is human generated A or B?};
    
    \draw[->] (input) -- (optA);
    \draw[->] (input) -- (optB);
    \draw[->] (optA) -- (judge);
    \draw[->] (optB) -- (judge);
    \draw[->] (judge) -- (output);
\end{tikzpicture}
\caption{LLM as a Judge. Binary human vs. LLM continuation detection.}
\label{fig:Judge}
\end{figure}
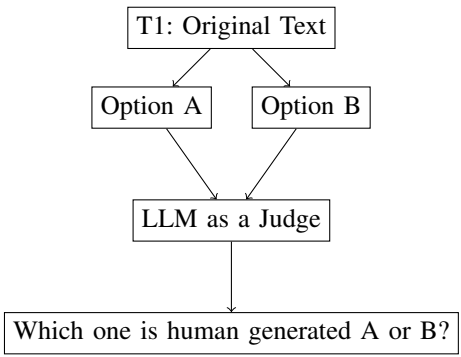

\paragraph{LLM-Based Prediction.} We prompt \textsc{GPT-4o} with T\textsubscript{1} and both candidate continuations and ask it to identify the continuation that best aligns stylistically and semantically with the original passage.

For transparency and reproducibility, we include the exact prompt used to elicit judgments from \textsc{GPT-4o}. The LLM is tasked with identifying the human-written
original continuation from 2 possible options. 

The following prompt template is used:The prompt asks the model to compare two possible continuations of a source text (\(T_1\)) and identify which continuation is human-authored.

Note that we randomize the order of the choices to prevent any bias from LLM as a judge. 

\subsubsection{Python Function}

The following function shows how the prompt is programmatically constructed:

\begin{lstlisting}[language=Python, caption={Prompt construction function.}]

def prompt_original_vs_llm(T1: str, A: str, B: str) -> str:
    """
    Returns a complete prompt asking the model to identify 
    which continuation (A or B) is the genuine 
    author-written follow-up to the source text T1.
    """
    return f"""You are an expert at detecting the style of written text
And identifying whether a text is written by a human or an AI language model.

You are given an original text T1 and two possible continuations A and B.
One of these is written by the original author of T1 and the other is 
generated by an LLM instructed to continue T1 in the same style.

Your task is to decide which of two continuations (A or B) is the genuine
author-written follow-up to a source text (T1).

Input
=====
T1:
{T1}

Continuation A:
{A}

Continuation B:
{B}

Answer with a single letter A or B.
"""
\end{lstlisting}

This prompt format is used consistently across all 600 instances in our evaluation, with T1, Continuation A, and Continuation B replaced with the appropriate text for each instance.

\section{Experiments}

\subsection{Experimental Design}

Our evaluation employs a paired comparison framework to directly assess both attribution approaches on identical data, enabling rigorous statistical analysis via McNemar's test.

\textbf{Dataset Structure.} We randomly sampled 600 instances from the HUMAN--AI PARALLEL CORPUS \cite{reinhart2024llms}, with 100 instances from each of six domains: academic, news, fiction, blogs, TV/movie scripts, and spoken transcripts. Each instance contains a 500-word human-authored prompt (T1), the original human continuation, and two LLM-generated continuations---one from GPT-4o and one from Llama-70B-Instruct. This structure yields 1,200 binary classification tasks: 600 human-vs-GPT comparisons and 600 human-vs-Llama comparisons.

\textbf{Randomization.} For each binary task, we randomly assigned the human and LLM continuations to positions A and B, preventing position bias. 

\textbf{Classifier Execution.} The Style Embedding baseline computed cosine similarities between T1 and each candidate, predicting the continuation with higher similarity as human-authored. The LLM Judge (GPT-4o) received the zero-shot prompt from Section III.C.3, returning predictions (A or B). Both systems operated independently on all 1,200 tasks.

\textbf{Analysis.} We computed accuracy for each classifier and applied McNemar's test to determine statistical significance of performance differences. Results were analyzed both in aggregate and segmented by domain and generator family (GPT vs. Llama) to identify systematic patterns in attribution quality.

\subsection{Metrics} 

We report the accuracy of both the baseline model and the LLM as a Judge Model. Given the task formulation involves binary classification between human- and LLM-generated continuations presented in a balanced 50/50 distribution, accuracy serves as an intuitive and meaningful metric \cite{sokolova2009systematic}. This setup ensures that a random classifier would achieve 50 per cent accuracy, providing a clear point of comparison for model performance.

To assess whether differences in performance between the two models are statistically significant, we apply McNemar’s Test—a non-parametric test commonly used for paired nominal data. 

We follow best practices in statistical evaluation of classifiers, using McNemar’s test as recommended in comparative studies of learning algorithms \cite{dietterich1998approximate}.

Since both models produce binary predictions over the same input instances, McNemar’s Test allows us to directly compare their error profiles by evaluating the number of discordant predictions. This test is particularly suited for our setting, where we aim to establish whether observed improvements are due to systematic differences rather than random variation.

\section{Results And Analysis}

\subsection{Performance Comparison}

The performance comparison is displayed in Tables I and Table II. 

\begin{table}[H]
\caption{Accuracy and McNemar test results comparing Base (Style Embeddings) vs LLM as a Judge (GPT-4o) performance across different segments for GPT-generated continuations}
\label{tab:gpt-mcnemar}
\centering
\begin{tabular}{@{}lccc@{}}
\hline
\textbf{Segment} & \textbf{Base} & \textbf{LLM} & \textbf{Winner} \\
\hline
Overall & 0.815 & 0.677 & Base$^\dagger$ \\
Fiction & 0.700 & 0.960 & LLM$^\dagger$ \\
Academic & 0.550 & 0.730 & LLM$^{**}$ \\
TVM & 0.950 & 0.680 & Base$^\dagger$ \\
News & 0.840 & 0.670 & Base$^{**}$ \\
Blog & 0.850 & 0.690 & Base$^*$ \\
Spoken & 1.000 & 0.330 & Base$^\dagger$ \\
\hline
\multicolumn{4}{l}{\scriptsize $^*p < 0.05$, $^{**}p < 0.01$, $^\dagger p < 0.001$}
\end{tabular}
\end{table}

\begin{table}[H]
\caption{Accuracy and McNemar test results comparing Base (Style Embeddings) vs LLM as a Judge (GPT-4o) performance across different segments for LLama-generated continuations}
\label{tab:llama-mcnemar}
\centering
\begin{tabular}{@{}lccc@{}}
\hline
\textbf{Segment} & \textbf{Base} & \textbf{LLM} & \textbf{Winner} \\
\hline
Overall & 0.805 & 0.847 & - \\
Fiction & 0.670 & 1.000 & LLM$^\dagger$ \\
Academic & 0.550 & 0.870 & LLM$^\dagger$ \\
TVM & 0.880 & 0.910 & - \\
News & 0.880 & 0.850 & - \\
Blog & 0.850 & 0.950 & LLM$^*$ \\
Spoken & 1.000 & 0.500 & Base$^\dagger$ \\
\hline
\multicolumn{4}{l}{\scriptsize $^*p < 0.05$, $^{**}p < 0.01$, $^\dagger p < 0.001$}
\end{tabular}
\end{table}

\subsection{Analysis}

The McNemar's test results reveal significant performance differences between the Base model (Style Embeddings) and LLM-based judge (GPT-4o) across various text domains. We analyze these findings for both GPT and Llama-generated continuations to understand the relative strengths of each approach.

\subsubsection{Comparative Performance Analysis}

\paragraph{Overall Performance} For GPT-generated continuations, the Base model significantly outperforms the LLM judge with an overall accuracy of 81.5\% versus 67.7\% ($p < 0.001$). However, for Llama-generated continuations, the LLM judge achieves slightly higher overall accuracy (84.7\% versus 80.5\%), but this difference is not statistically significant. This suggests that the Style Embeddings approach maintains more consistent performance across generators, while the LLM judge's effectiveness varies substantially depending on the source model. Interestingly GPT-4o performs worse when it comes when trying to distinguish between original author continuations and GPT-4o continuations.

\paragraph{Domain Specific Performance} Both approaches demonstrate pronounced domain specific strengths

\begin{itemize}
\item \textbf{Fiction}: The LLM judge significantly outperforms the Base model for both GPT (96\% vs. 70\%) and Llama (100\% vs. 67\%) generated continuations ($p < 0.001$). This represents the most substantial advantage for the LLM-based approach in any domain, suggesting that LLMs excel at evaluating narrative coherence and stylistic elements prevalent in fiction.
\item \textbf{Academic}: Similarly, the LLM judge shows significantly better performance in academic text evaluation for both GPT (73\% vs. 55\%, $p < 0.01$) and Llama (87\% vs. 55\%, $p < 0.001$) continuations. This may reflect LLMs' strong capabilities in assessing the structural and logical aspects characteristic of academic writing.
\item \textbf{TV/Movie}: The Base model significantly outperforms the LLM judge in the TV/Movie domain for GPT continuations (95\% vs. 68\%, $p < 0.001$). For Llama continuations, both models perform similarly (91\% vs. 88\%), with no significant difference. This suggests that style embeddings remain more effective for scripted dialogue, though the gap narrows in the Llama case.
\item \textbf{Spoken}: The most dramatic difference appears in spoken language, where the Base model achieves perfect accuracy (100\%) while the LLM judge performs poorly (33\% for GPT and 50\% for Llama continuations, $p < 0.001$). This indicates that style embeddings excel at identifying the distinctive markers of conversational speech that LLM judges might overlook or misinterpret.
\item \textbf{News and Blog}: For news content, the Base model significantly outperforms the LLM judge for GPT continuations (84\% vs. 67\%, $p < 0.01$), while neither approach shows a significant advantage for Llama continuations (88\% vs. 85\%). For blogs, the Base model is stronger for GPT continuations (85\% vs. 69\%, $p < 0.05$), but the LLM judge significantly outperforms for Llama continuations (95\% vs. 85\%, $p < 0.05$). These results suggest that attribution quality in informal domains may depend heavily on the generator family.
\end{itemize}

\subsubsection{Key Takeaways}

Our analysis reveals three principal findings:

\paragraph{1. Complementary Domain Strengths} The Base (Style Embeddings) model and LLM judge demonstrate complementary strengths across different text domains. LLMs excel at judging fiction and academic writing, while Style Embeddings show superior performance for spoken language, TV/movie dialogue, and certain news/blog cases. This complementarity suggests potential benefits from hybrid approaches that leverage the strengths of both methodologies.

\paragraph{2. Source Model Sensitivity} The performance gap between the two approaches varies substantially depending on whether the continuations are generated by GPT or Llama. The Base model maintains more consistent performance across GPT outputs, while the LLM judge shows stronger relative gains on Llama outputs. Notably, the LLM judge performs better overall when evaluating Llama-generated text (84.7\%) compared to GPT-generated text (67.7\%), despite both evaluations being performed by the same model (GPT-4o).

\paragraph{3. Structural versus Content Evaluation} The pattern of results suggests that Style Embeddings excel at capturing structural and register-specific linguistic features (particularly evident in spoken language and scripted dialogue), while LLM judges may prioritize semantic coherence and content quality (demonstrated by their superior performance in fiction, academic, and blog domains).

\subsubsection{Interpretation}

These performance differences likely stem from fundamental distinctions in how each approach conceptualizes and evaluates text quality:

Style Embeddings operate by learning distinctive linguistic patterns and statistical regularities specific to each domain. This approach appears particularly effective for domains with distinctive structural features, register-specific vocabulary, and consistent patterns (spoken language, TV dialogue). The perfect accuracy in the spoken domain suggests that conversational markers, and turn-taking structures create a distinctive signal that embeddings capture effectively.

In contrast, LLM-based judges employ a more holistic, semantically-rich evaluation approach that considers content coherence, narrative flow, and adherence to genre conventions. This appears advantageous when evaluating fiction, where plot consistency and character development may supersede purely stylistic features, and academic writing, where logical structure and argument quality are paramount.

The significant performance disparity when evaluating GPT versus Llama-generated continuations presents an intriguing phenomenon. One possible explanation is that LLM judges (using GPT-4o) may exhibit bias toward or against continuations from models in their own "family" or training lineage. In this case, it seems that GPT-4o is more likely to conflate the text that it has generated itself with human generated text, while it is able to better identify text produced by the LLama model.

These findings highlight the importance of selecting evaluation methodologies appropriate to the domain and generation model being assessed. They also suggest valuable directions for future work, including the development of hybrid evaluation approaches that combine style embeddings' strengths in structural pattern recognition with LLMs' capabilities in semantic and content evaluation.

\section{Conclusion}

This study positioned attribution quality as both a research challenge
and an evaluation benchmark, using the \textsc{Human–AI Parallel Corpus}
to systematically compare fixed style embeddings with an LLM judge.
By grounding hypothesis testing within a controlled benchmark, we
demonstrated how attribution mechanisms can be rigorously assessed
across domains and generator families, providing a replicable framework
for evaluating source-tracing methods.

We examined whether instruction-tuned LLMs can serve as effective judges
of authorship in a binary setting that pits a human continuation against
an LLM-generated rival. Contrary to our original hypothesis that an LLM
judge would match or exceed a stylistic baseline across the board, the
fixed style-embedding detector achieved the highest overall accuracy and
dominated in dialogue-rich and script genres. Nonetheless, the LLM
judge significantly outperformed the baseline in fiction and academic
domains, confirming that it captures complementary, content-level cues
that embeddings miss. McNemar analyses verified that these gains and
deficits are statistically reliable.

\textbf{Implications.}  The mixed outcome suggests that (i) present-day
LLMs possess only partial insight into their own generative
“fingerprint,” and (ii) provenance detection will benefit from hybrid
approaches that fuse structural style signals with the semantic
reasoning inherent to LLMs.

\textbf{Extensions of this work} should explore prompt engineering and chain-of-thought
rationales for LLM judges, fine-tune style encoders on
domain-specific corpora, and develop ensemble frameworks that
dynamically weight stylistic and semantic evidence. Such research will
move us closer to robust, genre-agnostic detectors of machine-generated
text—an essential safeguard as generative models continue to advance.

Ultimately, by treating attribution as a benchmarked evaluation problem,
this work contributes to the broader agenda of building transparent and
responsible attribution mechanisms for generative AI. The findings
highlight not only the limitations of current detectors but also the
promise of hybrid approaches.

Another interesting extension is to test attribution across multiple LLM families rather than the current human vs. single-LLM binary classification

\textbf{Reproducibility.} To support transparency and reuse, we release both the 
\textsc{Human–AI Parallel Detection Corpus} 
and the full source code for our experiments under the MIT license.

Our derived HUMAN–AI PARALLEL DETECTION CORPUS is built entirely on top of the HUMAN–AI PARALLEL CORPUS, which is released under the MIT license. As such, redistribution of derived splits and metadata under MIT is consistent with the original license. No third-party copyrighted material outside of this corpus is redistributed.

\section*{KEY LIMITATIONS AND FUTURE WORK}

\subsection*{LIMITATIONS}
\begin{itemize}
\item Limited to using only one model as a judge (GPT-4o) without exploring other LLMs, especially Reasoning models. 
\item Evaluation scale of 100 samples per domain is relatively modest for robust statistical analysis
\item No systematic ablation studies on prompt design, continuation length, or alternative embedding variants we performed. 
\end{itemize}

\subsection*{FUTURE WORK}
These are some concrete directions for future work to help address these limitations
\begin{itemize}
\item Evaluate multiple LLM judges to test generalizability beyond GPT-4o
\item Conduct ablation studies on prompt engineering and the fixed 500-word continuation length used in our setup
\item Repeat the study for larger samples for more statistically robust results
\end{itemize}

\section*{Authorship Statement}

This is solely the work of the author (s), relying on the cited papers for dataset and style embeddings as acknowledged throughout. Code generation tools were used for partially generating some code that went into this paper. Additionally, LLMs were also used in rephrasing parts of this paper but the original content is all human generated.

\bibliographystyle{IEEEtran}
\bibliography{anthology}

\appendices

\section{Worked Examples (Shortened)}

For clarity and space, we show only the first $\sim$50 words of each text, with ellipses indicating truncation. Full texts are available in the dataset release.

\subsection{Spoken Domain: Embedding Correct, LLM Incorrect}
\textit{Row 25. Gold = B (human).}

\paragraph{T\textsubscript{1} (prompt)}
\begin{quote}\footnotesize
Hello and welcome to the future of your ex podcast. I'm probably China's ux UI designer... We spoke about synthetic realities and how to apply generative design and AI to the process...
\end{quote}

\paragraph{Continuation A (GPT)}
\begin{quote}\footnotesize
In the circles of performers and educators alike... imagine harnessing the same principles in technology: taking raw data, structured algorithms... turning them into stories, movements, and vibrant hues...
\end{quote}

\paragraph{Continuation B (Human)}
\begin{quote}\footnotesize
There's a lot of dancers very interested in getting good at this... others say it’s not about what you do, but why... This can directly translate to discussions of digital art and design...
\end{quote}

\paragraph{Why Embedding Wins}
B retains spoken-register markers matching T1; A is polished and overly rhetorical.

---

\subsection{Academic Domain: LLM Correct, Embedding Incorrect}
\textit{Row 30. Gold = B (human).}

\paragraph{T\textsubscript{1} (prompt)}
\begin{quote}\footnotesize
Earth systems include the lithosphere, atmosphere, hydrosphere and biosphere... Of particular concern are anthropogenic influences on hazards such as earthquakes, landslides, floods, and subsidence...
\end{quote}

\paragraph{Continuation A (GPT)}
\begin{quote}\footnotesize
These transformative impacts occur through deforestation, agriculture, urban expansion... interactions between multiple hazards, referred to as multi-hazard scenarios... Consider, for example, groundwater abstraction and land subsidence...
\end{quote}

\paragraph{Continuation B (Human)}
\begin{quote}\footnotesize
Humans are important environmental agents... Anthropogenic processes influence natural hazard frequency or intensity... Examples include slope failures from vegetation removal, road construction, and inadequate drainage...
\end{quote}

\paragraph{Why LLM Wins}
B follows academic discourse conventions (citations, scoped definitions), while A drifts into rhetorical survey. The LLM judge tracks structure better than embeddings.

\section{Detailed McNemar's Test}

The detailed results of the McNemar's Test are shown in Table III and Table IV

\begin{table}[H]
\centering
\caption{McNemar results (GPT-generated continuations).}
\label{tab:mcnemar-gpt}
\setlength{\tabcolsep}{3pt}
\scriptsize
\resizebox{\columnwidth}{!}{%
\begin{tabular}{lccccc}
\toprule
Segment & Base Acc. & LLM Acc. & $p$-value & Sig. & Winner \\
\midrule
blog    & 0.850 & 0.690000 & 1.658900e-02 & True  & Base \\
tvm     & 0.950 & 0.680000 & 4.628673e-07 & True  & Base \\
fic     & 0.700 & 0.960000 & 2.556015e-06 & True  & LLM  \\
acad    & 0.550 & 0.730000 & 7.915897e-03 & True  & LLM  \\
news    & 0.840 & 0.670000 & 7.632079e-03 & True  & Base \\
spok    & 1.000 & 0.330000 & 1.355253e-20 & True  & Base \\
\midrule
Overall & 0.815 & 0.676667 & 1.574393e-07 & True  & Base \\
\bottomrule
\end{tabular}}
\end{table}

\begin{table}[H]
\centering
\caption{McNemar results (Llama-generated continuations).}
\label{tab:mcnemar-llama}
\setlength{\tabcolsep}{3pt}
\scriptsize
\resizebox{\columnwidth}{!}{%
\begin{tabular}{lccccc}
\toprule
Segment & Base Acc. & LLM Acc. & $p$-value & Sig. & Winner \\
\midrule
blog    & 0.850 & 0.950000 & 4.138947e-02 & True  & LLM  \\
tvm     & 0.880 & 0.910000 & 6.476059e-01 & False & ---  \\
fic     & 0.670 & 1.000000 & 2.328306e-10 & True  & LLM  \\
acad    & 0.550 & 0.870000 & 9.430375e-07 & True  & LLM  \\
news    & 0.880 & 0.850000 & 6.7776395e-01 & False & ---  \\
spok    & 1.000 & 0.500000 & 1.776357e-15 & True  & Base \\
\midrule
Overall & 0.805 & 0.846667 & 8.058114e-02 & False & ---  \\
\bottomrule
\end{tabular}}
\end{table}

\section{Code and Data Release}

\subsection{Dataset Release}
We release the \textsc{Human–AI Parallel Detection Corpus} on Hugging Face in this repository:
\url{https://huggingface.co/datasets/ephipi/human-ai-parallel-detection} under the MIT license. 
The dataset contains 600 balanced instances across six domains, each with a 500-word human prompt, a gold human continuation, and an LLM-generated continuation from either \textsc{GPT-4o} or \textsc{LLaMA-70B-Instruct}. \cite{touvron2023llama}
The release is designed to serve as an open benchmark for attribution research, enabling replication of our experiments and supporting further exploration of attribution quality assessment.

\subsection{Source Code Release}
All code required to reproduce our experiments is available in this public GitHub repository 
(\url{https://github.com/misamabbas/human-ai-parallel-detection/}), licensed under MIT. 
The repository includes data preprocessing scripts, style embedding baselines, LLM judge prompting utilities, and evaluation workflows (including statistical testing with McNemar’s test). 
We provide detailed documentation and example notebooks to facilitate reproducibility and to encourage extensions of our benchmark for new attribution methods or additional domains.

\end{document}